\documentclass{article}

\pdfoutput=1
\usepackage{cite}
\usepackage{arxiv}
\usepackage[utf8]{inputenc} 
\usepackage[T1]{fontenc}    
\usepackage{hyperref}       
\usepackage{url}            
\usepackage{booktabs}       
\usepackage{amsfonts}       
\usepackage{nicefrac}       
\usepackage{microtype}      
\usepackage{lipsum}
\RequirePackage{amsmath,amsfonts,amssymb}
\RequirePackage{graphicx,xcolor}

\title{Ensemble Clustering for Graphs: Comparisons and Applications}

\author{
  Valérie Poulin \\
  Tutte Institute for Mathematics and Computing \\
  Ottawa, Canada \\
  \texttt{vpoulin@gmail.com} \\
   \And
  François Théberge \\
  Tutte Institute for Mathematics and Computing \\
  Ottawa, Canada \\
  \texttt{theberge@ieee.org} \\
}

\begin{document}
\maketitle

\begin{abstract} 
We recently proposed a new ensemble clustering algorithm for graphs (ECG) based on the concept of consensus clustering. We validated our approach by replicating a 
study comparing graph clustering algorithms over benchmark graphs, showing that ECG outperforms the leading algorithms.
In this paper, we extend our comparison by considering a wider range of parameters for the benchmark, generating graphs with different properties.
We provide new experimental results showing that the ECG algorithm alleviates the well-known resolution limit issue, and that it leads to better stability of the partitions. 
We also illustrate how the ensemble obtained with ECG can be used 
to quantify the presence of community structure in the graph, and to 
zoom in on the sub-graph most closely associated with seed vertices.
Finally, we illustrate further applications of ECG by comparing it to previous results for community detection on weighted graphs, and community-aware anomaly detection.
\end{abstract}

\keywords{graph clustering \and ensemble \and consensus}

\section{Introduction}

\label{sec:1}
Most networks that arise in nature exhibit complex structure \cite{Girvan:2002,Newman:2003} with subsets of vertices densely interconnected
relative to the rest of the network, which we call communities or clusters.
Binary relational data-sets are typically represented as graphs $G=(V,E)$, where vertices $v \in V$
represent the entities, and edges $e \in E$ represent the relations between pairs of entities. 
Graph clustering aims at finding a partition of the vertices $V=C_1 \cup \ldots \cup C_l$ into good clusters.
This is an ill-posed problem \cite{Fortunato:2016}, as there is no universal definition of good clusters, leading to a wide variety of graph clustering 
algorithms \cite{Girvan:2002,Clauset:2004,Pons:2005,Newman:2006,Raghavan:2007,Reichardt:2006,Rosvall:2007,Blondel:2008},
with different objective functions.
In a recent study \cite{Yang:2016}, several state-of-the art algorithms implemented in the {\tt igraph} \cite{Csardi:2006} package were compared over a wide
range of artificial networks generated via the LFR benchmark \cite{Lancichinetti:2008}.
We recently introduced a new ensemble clustering algorithm for graphs (ECG), which compared favorably with leading algorithms from that study \cite{CNA_ECG:2019}.

The ECG algorithm is based on the concept of co-association consensus clustering. 
It is similar to other consensus clustering algorithms, in particular \cite{Lancichinetti:2012}, but differs in two major points: (1) the choice of an algorithm that alleviates the resolution limit issue for the generation step, and (2) the restriction to endpoints of edges for co-occurrences of vertex pairs, which keeps low computational
complexity.

The rest of the paper is organized as follows. We briefly describe the ECG algorithm in Section \ref{sec:2},
where we also recall some results from the previous comparison study. New results are included in the following three sections.

In Section \ref{sec:3}, we extend our study to a wider variety of graphs by varying the power law exponents of the LFR benchmark. 
Some of the advantages of ECG are its stability, and its ability to alleviate the well known resolution limit issue. We illustrate those properties in Section \ref{sec:4}. 
We also take a closer look at the edge weights generated by the ECG algorithm, showing that they can be good indicators of the presence (or absence) of community structure in a graph.
Applications are presented in Section \ref{sec:5}.
First, we consider a real graph, and show how ECG weights can be used to zoom-in on significant sub-graphs given some seed vertices. We then use ECG for two recently published applications, respectively for clustering {\it weighted graphs} \cite{CNA_Connes:2019}, and using graph clustering to find {\it anomalous nodes} \cite{CNA_CADA:2019}.
We wrap-up in the last section.

\section{Previous Results}
\label{sec:2}

Let $G=(V,E)$ be a graph where $V=\{1,2,\ldots,n\}$ is the set of vertices, and 
$E \subseteq \{(u,v) ~|~ u, v \in V,~ u < v\}$ is the set of edges. 
We consider undirected graphs.
Edges can have weights $w(e)>0$ for each $e \in E$. 
For un-weighted graphs, we let $w(e)=1$ for all $e \in E$.
The 2-core of a graph $G$ is its maximal subgraph whose vertices have degree at least 2.
Let $P_i = \{C_i^1,\ldots,C_i^{l_i}\}$ be a partition of $V$ of size $l_i$. We refer to the $C_i^j$
as {\it clusters} of vertices. We use $\mathbf{1}_{C_i^j}(v)$ to denote the indicator function for $v \in C_i^j$.

\subsection{The ECG Algorithm}

The ECG algorithm is a consensus clustering algorithm for graphs. Its generation step consists of independently obtaining $k$ randomized level-$1$ partitions from the multilevel-Louvain (ML) algorithm \cite{Blondel:2008}: ${\mathcal P} = \{P_1,\ldots,P_k\}$. Its integration step is performed by running ML on a re-weighted version of the initial graph $G=(V,E)$. The ECG weights are obtained through co-association. The weight of an edge $e=(u,v) \in E$ is defined as:

\begin{equation}
W_{\mathcal P}(u,v) = \left\{
\begin{array}{lc}
w_* + (1-w_*) \cdot \left(\frac{\sum_{i=1}^k \alpha_{{P}_i}(u,v)}{k}\right), & (u,v) \in \mbox{2-core of } G \\
w_*, & \mbox{otherwise}
\end{array}
\right.
\label{eq:w}
\end{equation}

\noindent where $0 < w_* < 1$ is some minimum weight and $\alpha_{P_i}(u,v) = \sum_{j=1}^{l_i} \mathbf{1}_{C_i^j}(u) \cdot \mathbf{1}_{C_i^j}(v)$ indicates if the vertices $u$ and $v$ co-occur in a cluster of $P_i$ or not. When running the ECG algorithm, the size $k$ of the ensemble and the minimum edge weight $w_*$
are the only parameters that need to be supplied. 
Guidelines for the parameters are given in \cite{CNA_ECG:2019}, where we also show that the results are not too sensitive with respect to those parameters.

\subsection{Comparison Study}

In \cite{CNA_ECG:2019}, we re-visited a recently published study of graph clustering algorithms,
comparing the best performing algorithms from that study with the ECG algorithm.
In general, we found ECG to yield better clusters with respect to all of the measures considered. Moreover, ECG generally found a number of communities much closer to the true value.

The algorithms are compared on graphs generated with the LFR benchmarks for 
undirected and unweighted graphs and with non-overlapping communities.
A key parameter when generating an LFR graph $G$ is the {\it mixing parameter} $\mu$, which sets the expected
proportion of edges in $G$ for which the two endpoints are in different communities.
We considered $.03 \le \mu \le .75$.  

\begin{figure}[th]
\centering
\includegraphics[width=12.25cm]{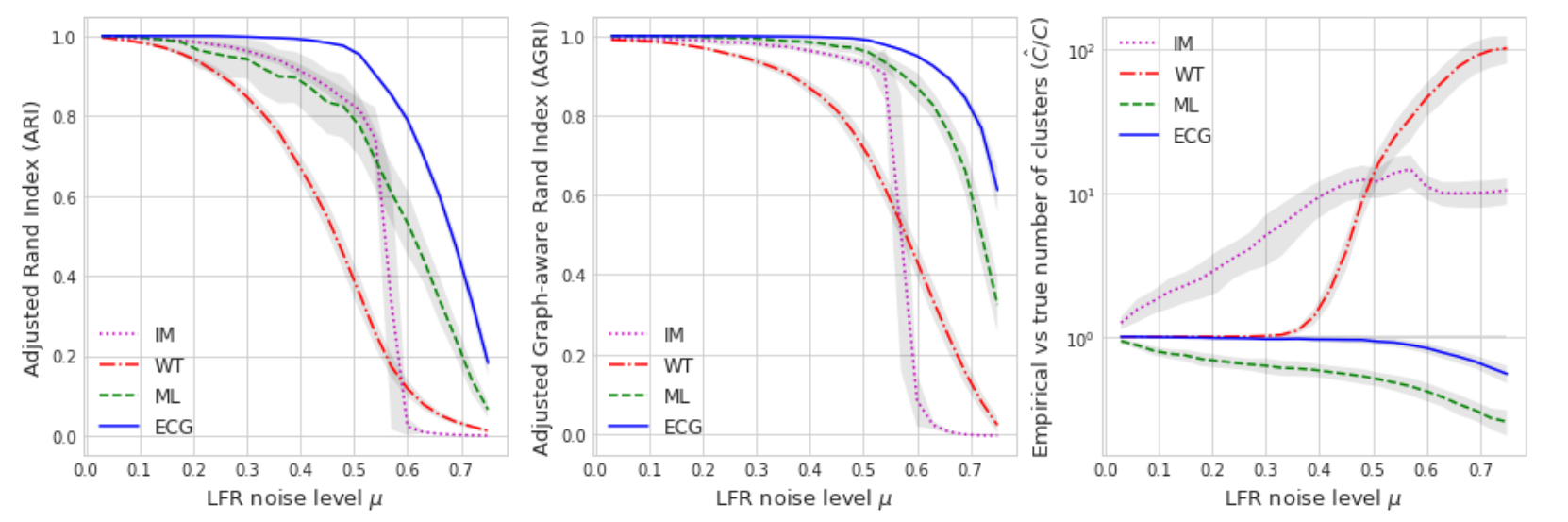}
\caption{
In the first two plots, we compare the accuracy of ECG with state-of-the-art algorithms: InfoMap (IM), WalkTrap (WT) and Louvain (ML). Results from each algorithm are compared with the true communities
for LFR graphs with $n=22,186$ vertices, and for various values of $\mu$, the proportion of noisy edges. For each value of $\mu$, we average over 10 LFR graphs; the shaded area shows the standard deviation.
We see that ECG outperforms all other algorithms.
In the third plot, we look at the ratio of the number of computed vs true communities.
We see that ECG remains very close to the desired value $\hat{C}/C=1$, as opposed to the other algorithms.
}
\label{fig:study}
\end{figure}

It was recently shown \cite{Poulin:2018} that graph-agnostic measures such as the 
adjusted RAND index (ARI) yield high scores for {\it refinements} of the true partition, while a graph-aware version (AGRI) gives high scores for {\it coarsenings} of the true partition when measuring graph partition similarities. It is thus recommended to
use both measures to compare algorithms, as we do throughout this paper.
We compared the true communities with those found by the ECG algorithm as well as three other state-of-the-art algorithms: InfoMap (IM) \cite{Rosvall:2007}, WalkTrap (WT) \cite{Pons:2005} and multilevel-Louvain (ML) \cite{Blondel:2008}.
The quality of the results from ECG are clear from the first two plots of Figure \ref{fig:study}, and the number of communities found with ECG remains much closer to the true number as the proportion of noise increases, as shown in the third plot.
Those conclusions are illustrative of the results we reported in \cite{CNA_ECG:2019}.

\section{Expanding the Comparison}
\label{sec:3}

In the LFR benchmark \cite{Lancichinetti:2009}, three important parameters are: the mixing parameter ($\mu$), the (negative) degree distribution power law exponent ($\gamma_1$), and the (negative) community size distribution power law exponent ($\gamma_2$). It is generally recommended to use $2 \le \gamma_1 \le 3$ and $1 \le \gamma_2 \le 2$ to model realistic networks \cite{Lancichinetti:2009}, \cite{Barabasi:2016}.
In the previous section, we compared ECG with other state-of-the-art algorithms with the same parameter choices as in \cite{Yang:2016}. While we considered a wide range for parameter $\mu$, the power law exponents were fixed at $\gamma_1=2$ and $\gamma_2=1$.
In this section, we summarize the impact of those parameters on the types of networks that are generated, and we re-visit the comparison results, exploring a wider set of graphs.

\begin{figure}[ht]
\centering
\includegraphics[width=8cm]{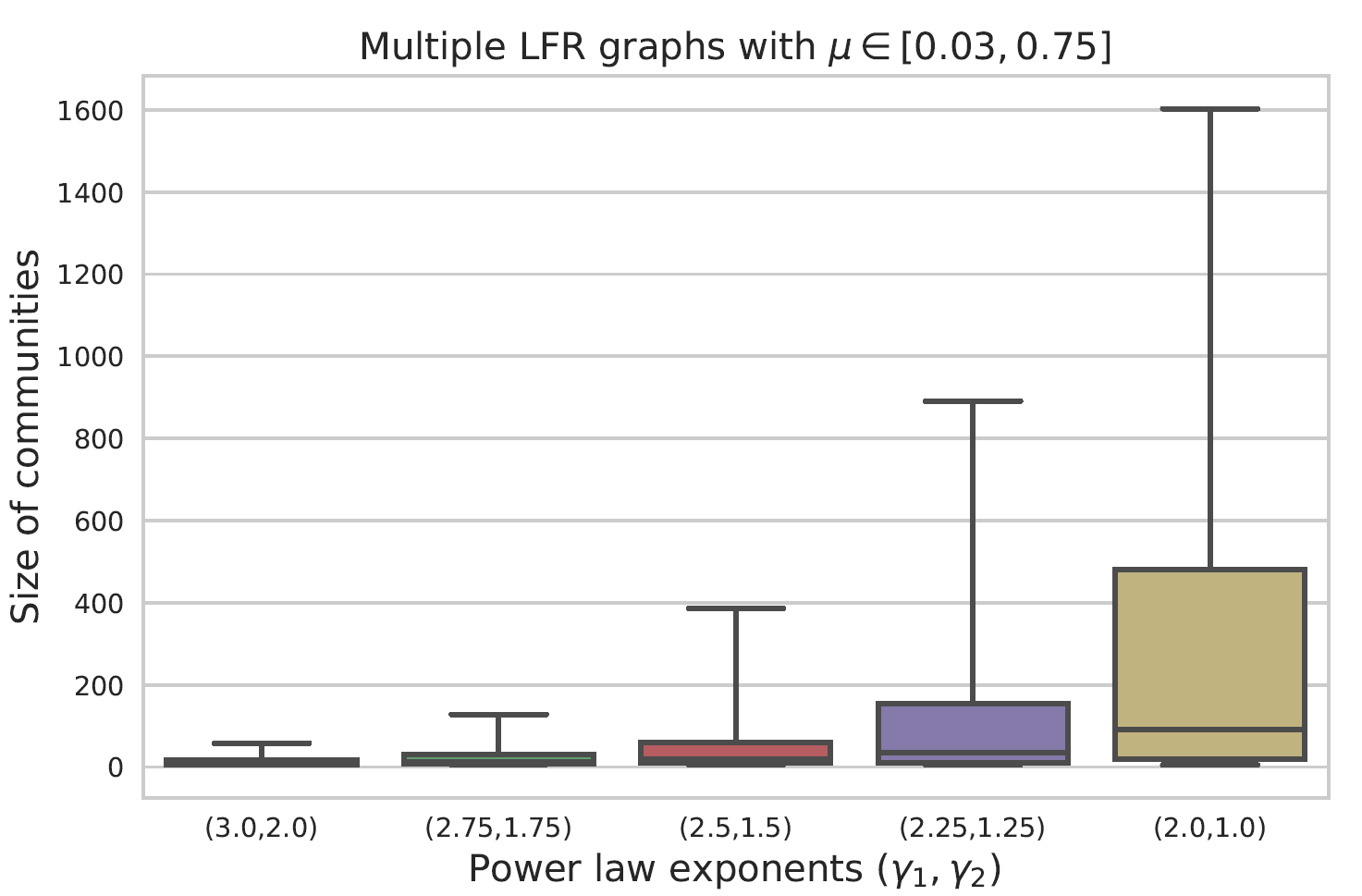}
\caption{
We selected 5 choices for the power law parameters $(\gamma_1,\gamma_2)$ which are representative of various types of networks obtained with the LFR benchmark, and we look at the distribution of the sizes of communities. We see that with the largest recommended values $(\gamma_1,\gamma_2) = (3,2)$, we get small communities of homogeneous size. 
As the exponents decrease, the sizes of the communities get more
heterogeneous. All results were obtained by averaging over
10 graphs with 22,186 nodes for every choice of parameters $(\mu, \gamma_1,\gamma_2)$.
}
\label{fig:LFRa}
\end{figure}

In Figure \ref{fig:LFRa}, we show some topological graph differences over 5 choices of parameters $(\gamma_1,\gamma_2)$ in the recommended range. We see that for larger values of those parameters, the communities generated are small and of similar size while
smaller values for $(\gamma_1,\gamma_2)$ yield graphs with more heterogeneous community sizes, which are more realistic.

\begin{figure}[ht]
\centering
\includegraphics[width=12.25cm]{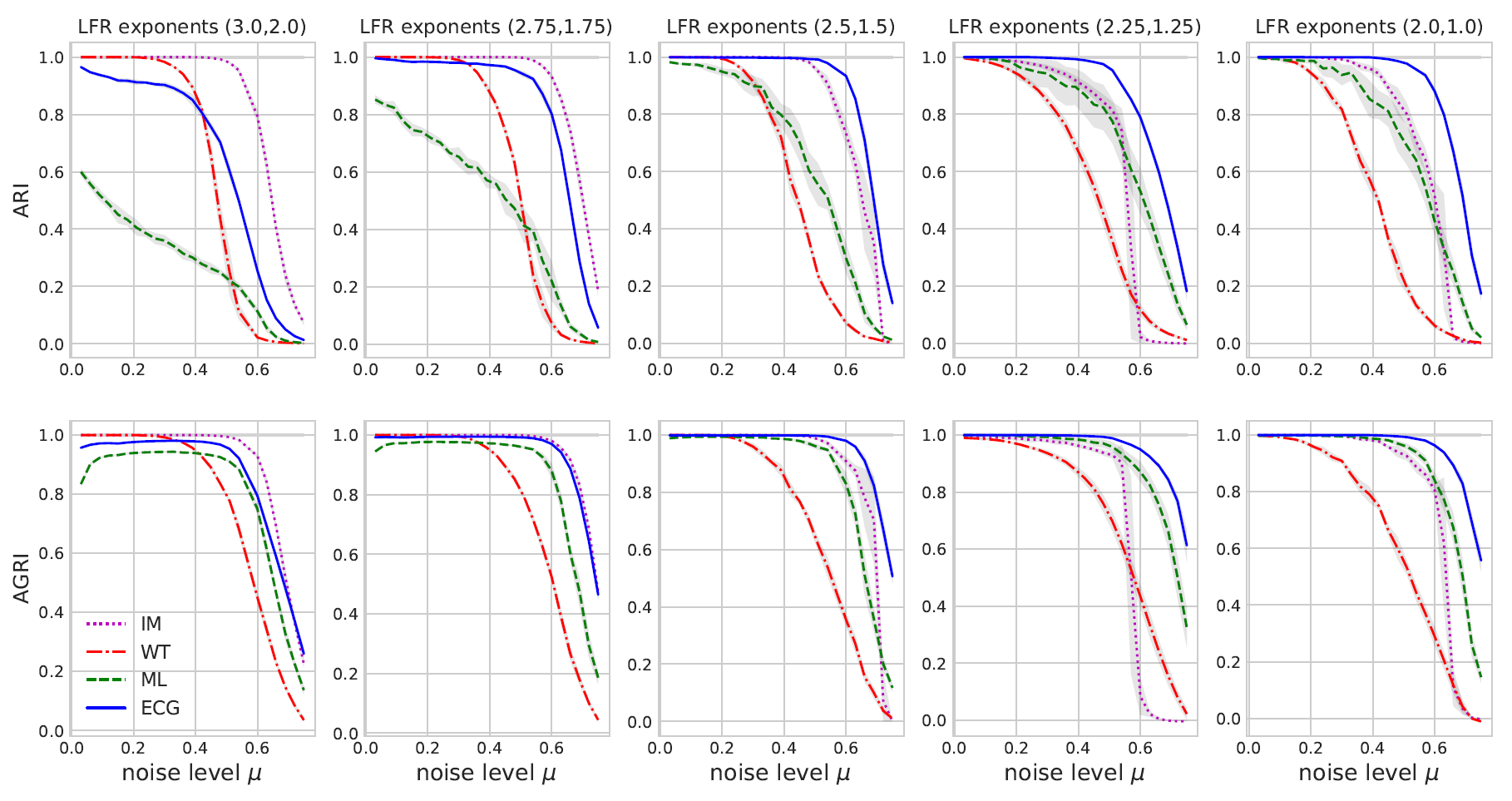}
\caption{We measure the quality of the communities found by the InfoMap (IM), WalkTrap (WT), Louvain (ML) and ECG algorithms over LFR graphs with 5 different choice of power law exponents. For the plots on the left, we see that the Louvain algorithms does not do very well, ECG does much better and the InfoMap yields the best results. Recall that those graphs have many small communities of similar sizes. As we move toward the right, the Louvain algorithm does better and ECG yields the best results. Those graphs have more heterogeneous community sizes.
}
\label{fig:LFRb}
\end{figure}

In Figure \ref{fig:LFRb}, we again compare ECG with IM, WT and ML. 
For the larger values of $(\gamma_1,\gamma_2)$, we see that the ML algorithm does not do very well, with ECG doing much better and IM yielding the best results. As before, we use both the ARI measure and its graph-aware counterpart AGRI. As the exponents decrease, indicative of more heterogeneous community size distribution, 
we see that the ML algorithm does better, and ECG gives the best results overall. 

Therefore, by expanding the comparison over a wider range of LFR parameters, we see that ECG generally gives better results, with the exception of graphs with small communities of homogeneous size, where IM is slightly better.

\section{Resolution Limit and Stability}
\label{sec:4}

At the heart of ECG is the fact that we use
multiple runs of the {\it single-level} Louvain algorithm 
to build an ensemble of {\it weak} (or local) partitionings of the vertices. In this section, we illustrate the two main reasons for this choice.

\subsection{Resolution Issue: Ring of Cliques Illustration}

The resolution limit issue is well illustrated by the infamous ring of cliques
example, where the $n$ vertices form $l$ cliques (full sub-graphs) of size $m$, wired together as a ring. For some choices of $l$ and $m$, grouping pairs of adjacent cliques yields a higher modularity value than the natural choice of each clique forming its own cluster \cite{Fortunato:2007}. The latter yields higher modularity if and only if $m(m-1) > l-2$. 
In \cite{CNA_ECG:2019}, we show that choosing a small value for $w_*$ in (\ref{eq:w}) can alleviate this issue. In particular, choosing $w_*<1/n$ avoids the issue altogether.

\begin{figure}[th]
\centering
\includegraphics[width=12.25cm]{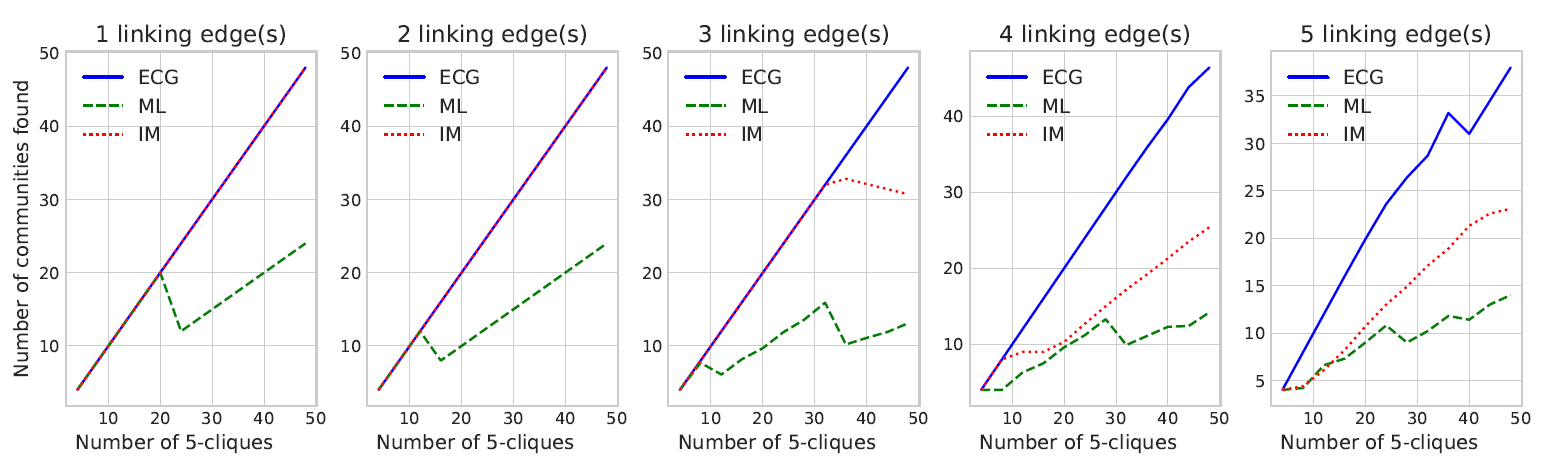}
\caption{
In each plot, we consider $l$ cliques of size $m=5$ where contiguous cliques are linked by 1 to 5 edges, respectively. We compare the number of communities found by the InfoMap (IM), Louvain (ML) and ECG algorithms. The resolution limit phenomenon is clearly seen with the ML algorithm. The IM algorithm fails to find the right number of communities when we increase the number of edges between the cliques, while ECG remains more stable.
}
\label{fig:roc1}
\end{figure}

In Figure \ref{fig:roc1}, we look at rings of $l$ cliques of size $m=5$, with 1 to 5 edges between contiguous cliques. 
For the ML algorithm, we see the resolution limit issue when $l>20$ (with 1 edge between contiguous cliques), which agrees with the known results. The IM algorithm is stable when only a few edges link the cliques, but quickly becomes unstable as more edges are added, while the ECG algorithm remains very stable keeping the default choice of $w_* = .05$.

\begin{figure}[ht]
\centering
\includegraphics[width=12.25cm]{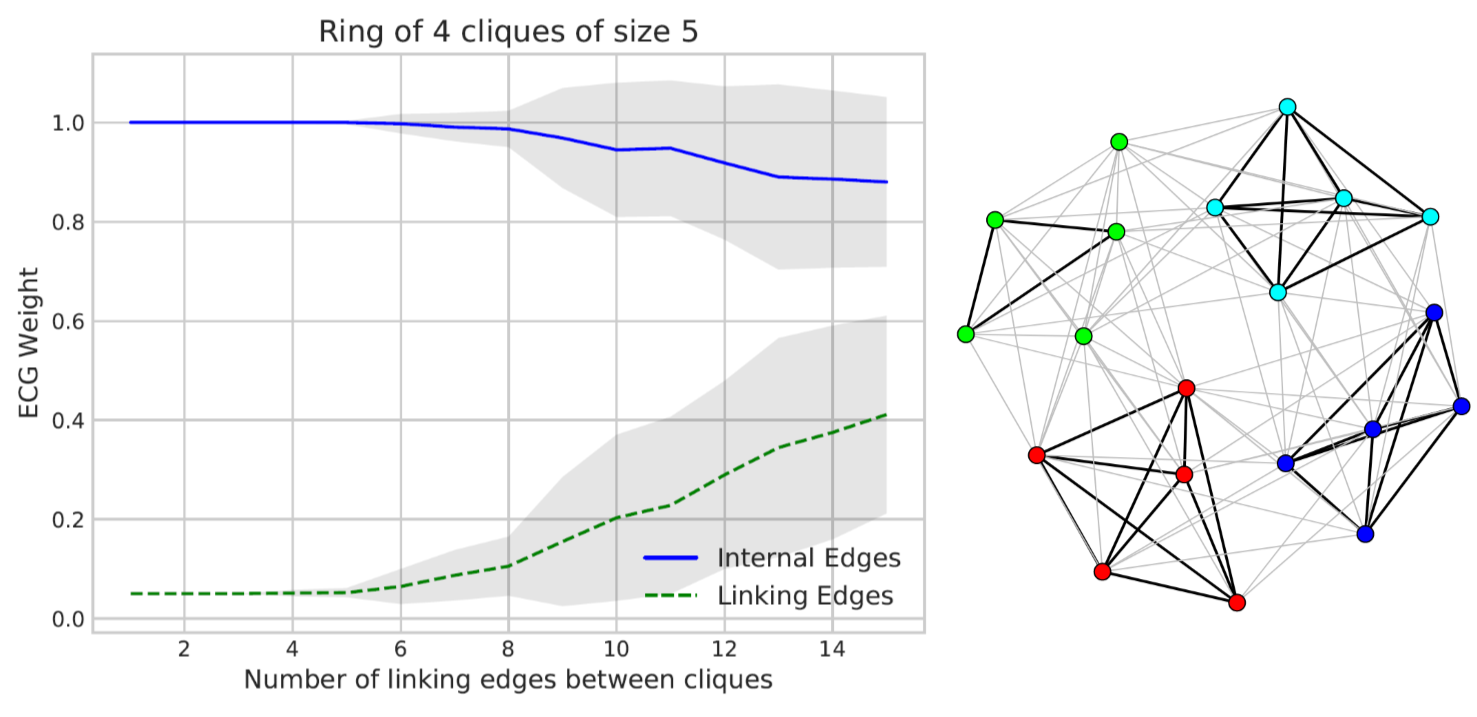}
\caption{
We add 1 to 15 edges between contiguous cliques in a ring of 4 cliques of size 5, and we look at the effect on the ECG edge weights for edges internal to the cliques, or external edges linking the contiguous cliques. In the right plot, we look at the case with 15 edges between cliques; thick edges are the ones where the ECG weight is 0.8 or above.
}
\label{fig:roc2}
\end{figure}

We further illustrate this stability in Figure \ref{fig:roc2}, where we add up to 15 edges between the cliques of size 5 in a ring with 4 cliques.
We see that even when the number of edges linking the cliques is comparable to the number of edges within each clique, the signal obtained with the ECG weights still favours the cliques. This behaviour allow to better identify communities in noisy graphs.
In the right plot of Figure \ref{fig:roc2}, we show the case where 15 edges are added between
contiguous cliques. Thicker edges are the ones where the ECG weights are above 0.8. 
We see that most of the clique structure is still 
captured when looking only at those high weight edges.

\subsection{Weight distribution and community structure}

\begin{figure}[th]
\centering
\includegraphics[width=12.25cm]{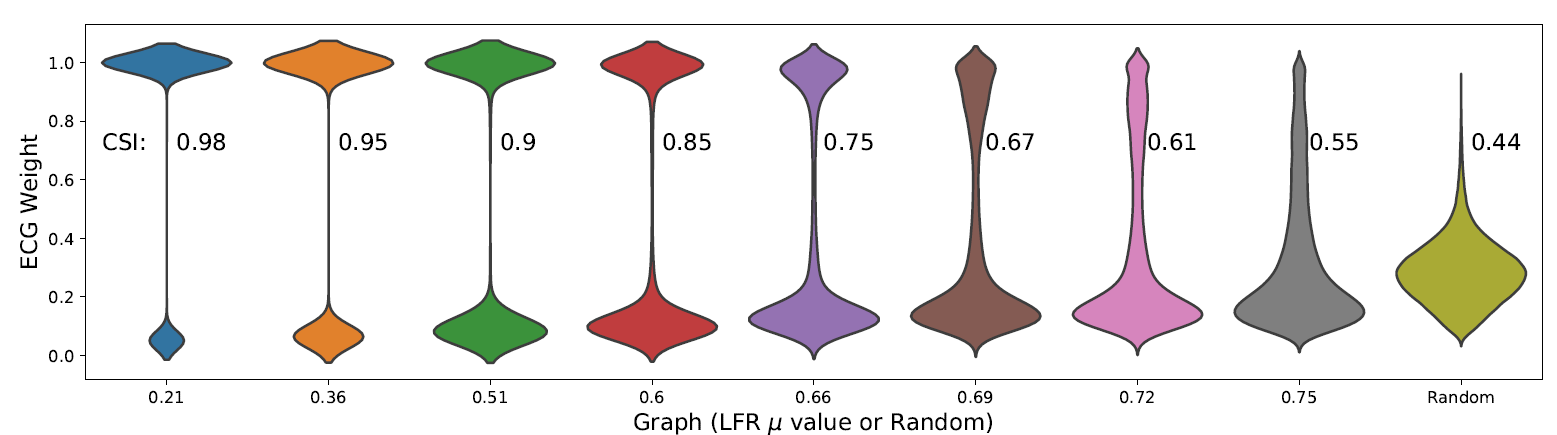}
\caption{
Violin plots of the ECG weight distribution for a family of LFR graphs with $n=22,186$ nodes, parameters $\gamma_1=2$, $\gamma_2=1$ and $.21 \le \mu \le .75$. We also compare with a random graph of the same size and degree distribution as the graph with $\mu = 0.21$.
We see the bi-modal distribution over LFR graphs up to a very high noise level. For large $\mu$, the signal gets weaker. It is even weaker for the random graph. The Community Strength Indicator (CSI) is also reported.
}
\label{fig:violin}
\end{figure}

We compare the ECG weight distribution over LFR graphs where we vary the mixing parameter. We also compare with a random graph having the same degree distribution as one of the LFR graphs. 
Bi-modal distribution 
of the ECG weights near the boundaries (0 and 1) is indicative of
strong community structure. We propose a simple community strength indicator (CSI) based on the point-mass 
Wasserstein distance. For all edges $(u,v) \in E$, with $W_{\mathcal P}(u,v)$ from (\ref{eq:w}), we define:

\begin{equation}
CSI = 1 - 2 \cdot \frac{1}{|E|} \sum_{(u,v) \in E} \min \left( W_{\mathcal P}(u,v), 1-W_{\mathcal P}(u,v) \right)
\label{eq:wd}
\end{equation}

such that $0 \le CSI \le 1$, where a value close to 1 is indicative of strong community structure, random weights $W_{\mathcal P}(u,v)$ yield a value close to 0.5, and $CSI=0$ when all $W_{\mathcal P}(u,v)=0.5$.
In Figure \ref{fig:violin}, we see the bi-modal distribution of the weights for low and mid-range choices of $\mu$, along with high $CSI$ values. 
For larger values of $\mu$, the distribution is not as clear, and there are less and less edges with weight close to 1, which indicate a weak community structure, as confirmed by the $CSI$ values. 
The random graphs have low weights only, which is indicative of the absence of community structure. This example illustrates how
the distribution of edge weights obtained with ECG, along with the proposed $CSI$, can be used to assess the strength of community structure in a graph.

\subsection{Stability of ECG}

So far, we saw that the ECG weights are useful to alleviate the resolution issues of modularity, and can also be used to assess the presence of community structure in a graph. 
We illustrate another advantage of ECG which is to significantly reduce the instability in the ML algorithm. 
To test for stability, we run the same algorithm twice on each graph considered, and we compare the two partitions obtained with the ARI (or AGRI) measure. 
 
\begin{figure}[th]
\centering
\includegraphics[width=12.25cm]{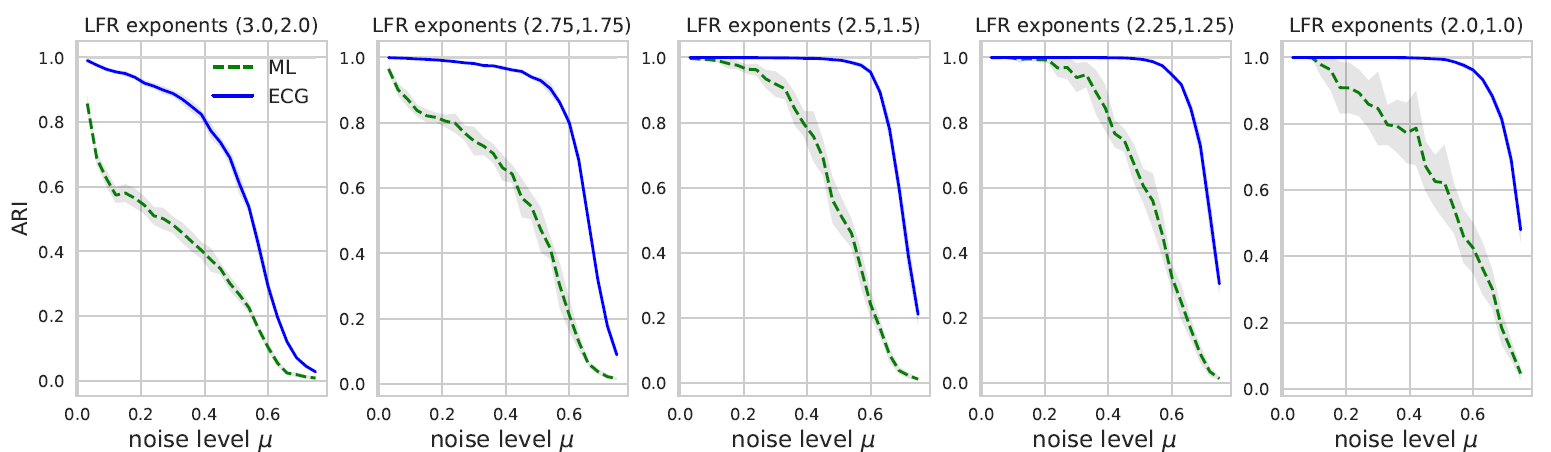}
\caption{
We compare the stability of the communities found by the Louvain (ML) and ECG algorithms over LFR graphs with 5 different choice of power law exponents.
Partitions obtained in distinct runs for each algorithm are compared via the ARI measure. We see the much improved stability with ECG. Conclusions are the same with the AGRI measure (not shown).}
\label{fig:stability}
\end{figure}

In Figure \ref{fig:stability}, we did this for the ML and ECG
algorithms over LFR graphs with the same parameters as in the previous section. We see that in all cases, ECG greatly improves the stability of the Louvain algorithm.

\section{Other Applications of ECG}
\label{sec:5}

In this Section, we look at a few applications with ECG.

\subsection{ECG weights and a dimmer process}

Assume that we are interested in some {\it seed} vertices in a graph. 
In large graphs, it is not clear how to properly ``zoom in'' on the sub-graph
showing the main interactions around the seed vertices.
Taking the seed's ego-nets (immediate neighbours) may not show all the strong interactions, and taking the entire clusters from a partition which contain the seed vertices may be too large. The weights provided by ECG can be used to define a dimmer-like process around the seed vertices, thus highlighting the sub-graphs that are the most tightly connected to the seeds.

Consider a graph $G$, a seed vertex $v$ and $G_v \subset G$ the sub-graph of $G$ formed by keeping only the ECG cluster containing vertex $v$. Given some threshold $\theta$, we delete all edges in $G_v$ with ECG weights below $\theta$,  and we keep the connected component sub-graph containing vertex $v$.
Increasing $\theta$ from 0 to 1 provides a hierarchy of sub-graphs of decreasing size which all contain vertex $v$.

\begin{figure}[th]
\centering
\includegraphics[width=12.25cm]{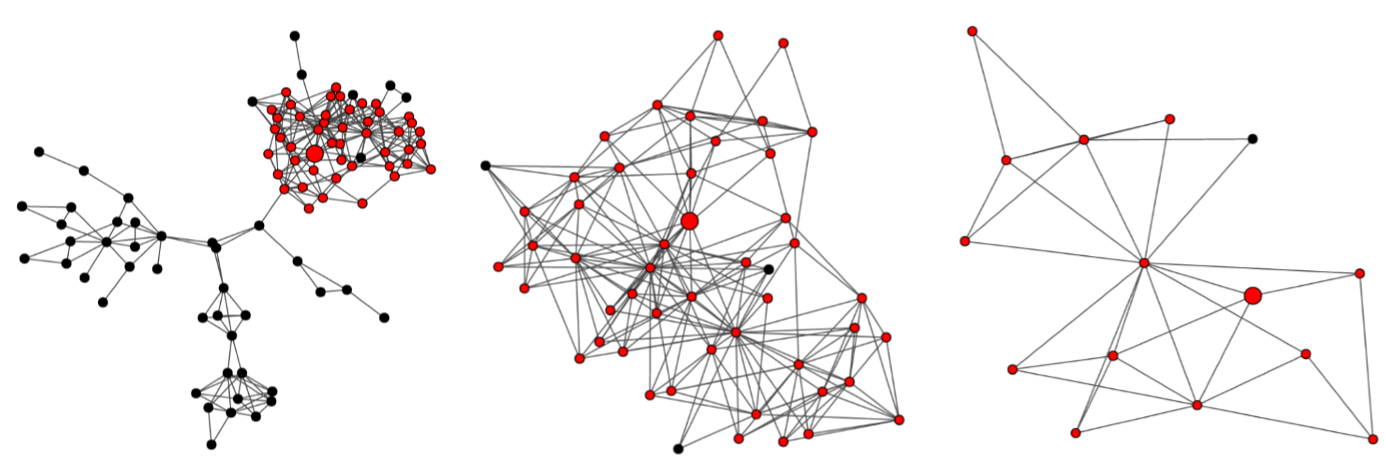}
\caption{
We consider a seed vertex (display with larger size) from the Amazon co-purchasing graph. Vertices from the same ground truth communities are displayed in red, and other vertices are displayed in black. From left to right, we display respectively (i) the entire sub-graph obtained from the ECG part that contains the seed vertex, (ii) a connected sub-graph with ECG edge weights above 0.1 containing the seed vertex, and (iii) a connected sub-graph with ECG edge weights above 0.72 containing the seed vertex. While the first plot has many spurious vertices, as we zoom in, most nodes we retain are in the same true community as the seed node.
}
\label{fig:amazon}
\end{figure}

As an illustration of this process, we consider the Amazon co-purchasing graph available from the SNAP repository \cite{snapnets}. This graph has 334,863 nodes and 925,872 edges.
There are over 75,000 communities, 5000 of which are identified as the top ones. 
We picked a vertex $v$ that belongs to one of those top communities\footnote{vertex {\it 112067} in the minimized data from \cite{snapnets}.}. We ran ECG, and isolated the sub-graph $G_v$ induced by the vertices in the ECG cluster that contains $v$. In Figure \ref{fig:amazon}, we gradually increase the threshold $\theta$, keeping only edges in $G_v$ with ECG weight above that threshold, and showing the connected component containing $v$. In the first plot, we set $\theta=0$, thus showing $G_v$ ($v$ is shown with larger size). Vertices in red belong to the same ground truth community as $v$.
While we see a lot of spurious vertices in the first plot, discarding edges with low ECG weights (setting $\theta=0.1$) yields the second sub-graph, where all ground truth vertices are retained.
The last plot shows a more aggressive filtering, where we retain only edges with high ECG weights (setting $\theta=0.72$). This reveals a tightly connected subset of vertices around the seed vertex $v$.

\subsection{Application to weighted graphs}

So far, all of our comparisons for ECG were done over un-weighted
graphs. In \cite{CNA_Connes:2019}, among other things, the authors study various edge re-weighting schemes and graph clustering algorithms over {\it weighted} LFR graphs. They found that using the GloVe re-weighting function along with the Label Propagation (LP) algorithm \cite{Raghavan:2007} gave the best results for identifying communities. 

\begin{figure}[th]
\centering
\includegraphics[width=12.25cm]{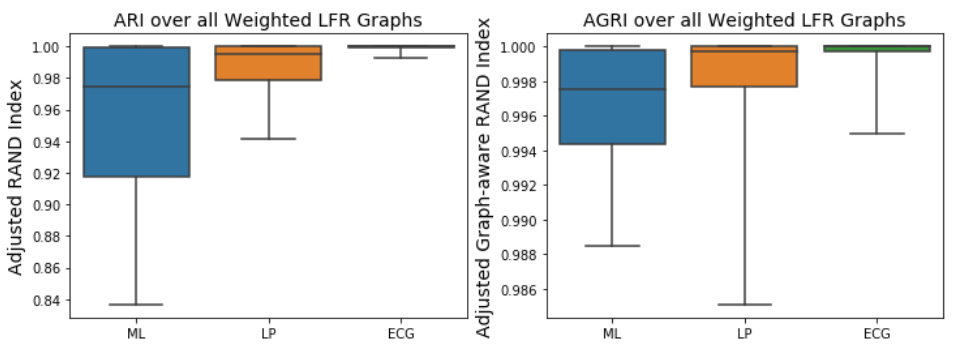}
\caption{
Results over weighted LFR graphs for 3 graph clustering algorithms: Multilevel-Louvain (ML), Label Propagation (LP) and ECG. Results are obtained over all graphs in \cite{Dugue:Code} for which the LP algorithm converged. GloVe re-weighting as reported in \cite{CNA_Connes:2019} is applied to all graphs.
}
\label{fig:connes}
\end{figure}

We re-created this experiment, using the same graphs available at \cite{Dugue:Code}, and the same re-weighting function with the
best choice of parameters as reported in Table 1 of \cite{CNA_Connes:2019}. 
We compared LP, ML and ECG algorithms. 
While we generally obtained good results with LP, 
we had to discard some runs as this algorithm sometimes 
failed to converge. 
We show our results in Figure \ref{fig:connes}, where we summarize the ARI and AGRI scores we obtained over all graphs for which the LP algorithm converged. We see that the
results are better in general with ECG, and with much improved stability.

\subsection{Community-aware anomaly detection}

In \cite{CNA_CADA:2019}, the authors propose $CADA$, a community-aware method for detecting anomalous vertices. In a nutshell,
for each vertex $v \in V$, let $N(v)$ represent the number of neighbors of $v$, and $N_c(v)$ the number of neighbors of $v$ that belong to the most represented community obtained with the IM or ML algorithm.
They define:
$CADA_x(v) = \frac{N(v)}{N_c(v)}$ where $x \in \{IM, ML\}$ indicates the clustering algorithm used.
They compare their algorithm to other methods by generating LFR graphs with degree exponent $\gamma_1=3$ and community size exponent $\gamma_2=2$. As we saw earlier, this choice corresponds to small communities of homogeneous size, where the IM algorithm performs best. 
We re-visited this approach with ECG, considering different values for the power law exponents, as in section ``Expanding the Comparison''. 
We generated LFR graphs with $n=22,186$ nodes and various values for the mixing parameters. For each graph, we introduced 200 random anomalous nodes with the same degree distribution, as in Figure 1 of \cite{CNA_CADA:2019}.

\begin{figure}[th]
\centering
\includegraphics[width=12.25cm]{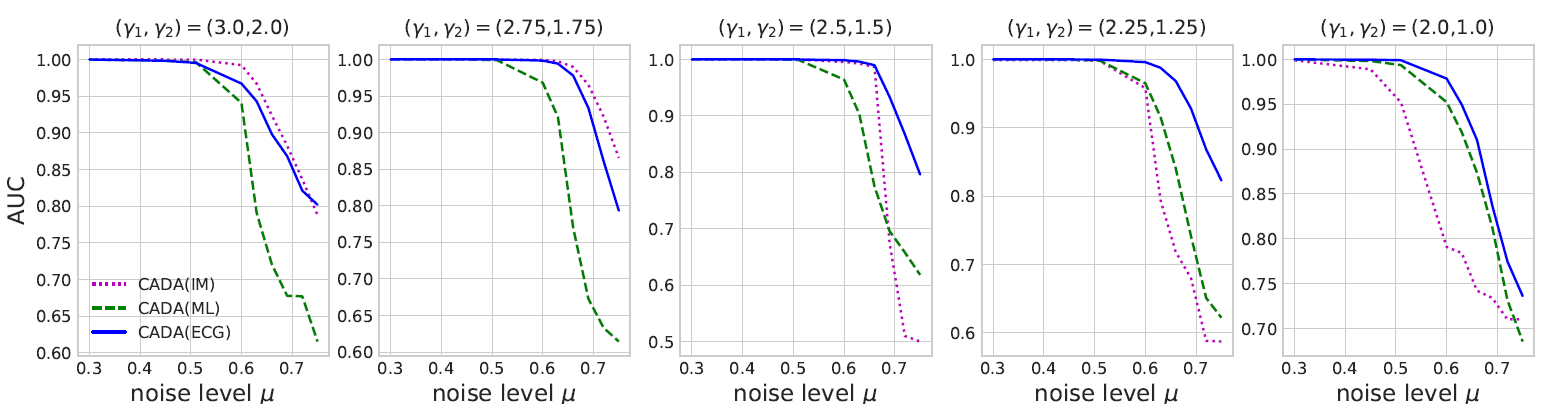}
\caption{We compare three flavours of the CADA algorithm, using the InfoMap (IM), Louvain (ML) and ECG. For each value of $.3 \le \mu \le .75$, we generated 10 LFR graphs of size 22,186, along with 200 random anomalous nodes with the same degree distribution. We considered 5 different choices for the LFR power law exponents. Results are compared via the area under the ROC curve (AUC).
}
\label{fig:CADA}
\end{figure}

In Figure \ref{fig:CADA}, we compare $CADA_{ECG}$ with $CADA_{IM}$ and $CADA_{ML}$
using the areas under the ROC curves (AUC). 
We see that for large choices of the power law exponents, the IM version does best.
This is the only choice of parameters used in \cite{CNA_CADA:2019}. 
As we decrease the values of the exponents, we see that using ECG becomes a better choice, in particular for large values of $\mu$. This is due to the increased stability and the ability to distinguish the signal from the noise provided by the ECG weights, which we illustrated earlier in section ``Resolution Limit and Stability''.

\section{Conclusion}
\label{sec:6}

In \cite{CNA_ECG:2019}, we proposed ECG, a new graph clustering algorithm based on the concept of consensus clustering, and we compared it to other algorithms by re-creating the study in \cite{Yang:2016}.
In this paper, we compared ECG with state-of-the-art algorihms over a wider range of graphs, showing ECG to be the best performing algorithm in most cases. We provided empirical evidence for the two main advantages of ECG: its ability to greatly reduce the resolution limit issue of modularity, and its high stability. We also illustrated how the edge weights generated in ECG can be used to assess the presence of community structure in graphs. 
Finally, we favourably applied ECG to three tasks: we showed how to extract relevant sub-graphs around seed vertices, we used ECG to find communities in {\it weighted} graphs, and we applied ECG for the task of detecting anomalous vertices in graphs.

\bibliographystyle{unsrt}
\bibliography{main}

\end{document}